\documentclass[letterpaper]{article} 
\usepackage{aaai2027} 
\usepackage[hyphens]{url}  
\usepackage{graphicx} 
\usepackage{natbib}  
\usepackage{caption} 
\usepackage{algorithm}
\usepackage{algorithmic}
\usepackage{amsmath}
\usepackage{amssymb}
\usepackage{tabularx}
\usepackage{newfloat}
\usepackage{listings}
\DeclareCaptionStyle{ruled}{labelfont=normalfont,labelsep=colon,strut=off} 
\floatstyle{ruled}
\newfloat{listing}{tb}{lst}{}
\floatname{listing}{Listing}

\usepackage{booktabs}

\title{Metaplasticity as adaptive gradient preconditioning for incremental learning}

\author{
    Isabelle Aguilar\corresponding\textsuperscript{\rm 1},
    Zayn Andre Zainal\textsuperscript{\rm 1},
    Omid Kavehei\textsuperscript{\rm 1},
}
\affiliations{
    \textsuperscript{\rm 1}School of Biomedical Engineering, The University of Sydney\
    Maze Cres, Darlington, NSW, Australia, 2008\\
    \{isabelle.aguilar, andre.zainal, omid.kavehei\}@sydney.edu.au,
}

\begin{document}

\maketitle

\begin{abstract}
Biological intelligence naturally prevents catastrophic forgetting through Complementary Learning Systems (CLS) theory, a macroscopic consolidation process driven at the local level by synaptic metaplasticity: the continuous, history-dependent neuromodulation of individual synapses. While artificial neural networks struggle with the stability-plasticity dilemma in non-stationary environments, existing solutions often require task labels or incur massive memory overhead, diverging from biological reality. Re-framing this localized neuromodulation as an optimization-driven process, we introduce \textbf{SynGAP}: \textbf{Syn}aptic \textbf{G}eometric \textbf{A}daptive \textbf{P}reconditioning. SynGAP is a task-free continual learning framework based on adaptive gradient preconditioning. Rather than relying on explicit episodic triggers, SynGAP simulates real-time metaplasticity by maintaining an exponential moving average of the Fisher Information Matrix over a continuous data stream. During the optimization step, these dynamic metaplastic states are translated into a bounded multiplicative mask that preconditions raw gradients, selectively attenuating updates to critical historical parameters. Empirical evaluations demonstrate SynGAP's superior ability to mitigate catastrophic forgetting compared to established baselines. On the Split CIFAR-100 benchmark, SynGAP delivers a $4\times$ increase in accuracy compared to EWC++ and outperforms Experience Replay (ER) by almost $10\%$, while reducing the forgetting measure by over $10\%$ against both methods. Furthermore, on the CORe50 benchmark, SynGAP achieves about $68\%$, a $10\%$ improvement over optimizer baselines. By mathematically formalizing continuous biological metaplasticity as stable gradient-based regularization, SynGAP offers a highly robust and memory-efficient solution for adaptive intelligence at the edge.
\end{abstract}

\section{Introduction}
\label{sec:intro}
\begin{figure}[htpb!]
\centering
\includegraphics[width=0.50\textwidth]{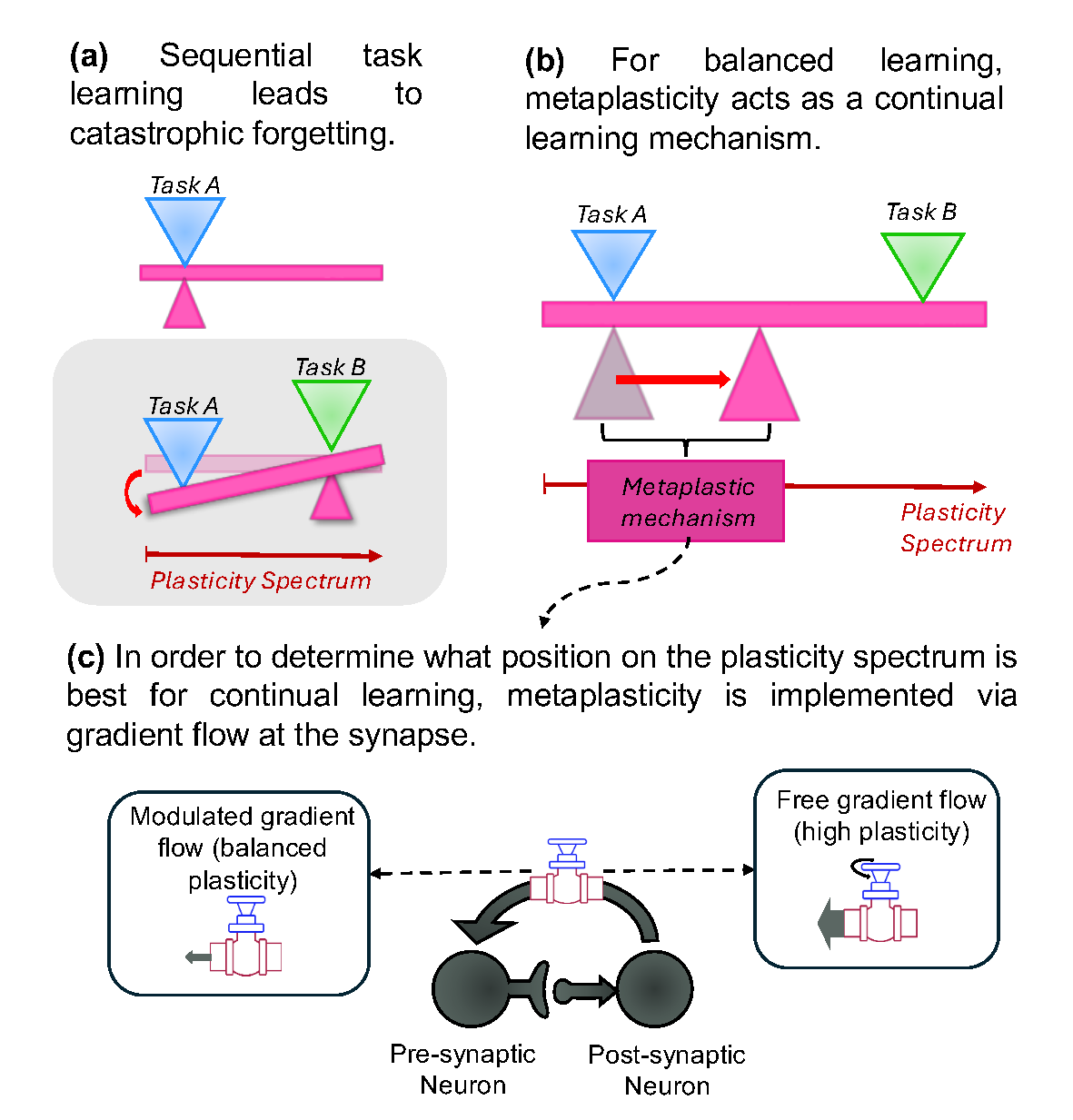}
\caption{\textbf{Balancing the stability-plasticity dilemma requires metaplastic control.} \textbf{(a)} Conventional learning leads to catastrophic forgetting of an old task (Task A), where the introduction of a new task (Task B) causes an unstable balance on the plasticity spectrum. \textbf{(b)} To counteract this, a metaplasticity mechanism shifts the "fulcrum" of learning on a seesaw that holds both tasks, balancing the consolidation of old knowledge with the intake of new information. \textbf{(c)} At the network level, this is implemented via localized gradient flow modulation at the synapse. As an adaptive valve, metaplasticity regulates gradient updates to determine the most balanced position along the plasticity spectrum.}
\label{fig:fig1} 
\end{figure}

Biological intelligence is grounded in its ability to learn continuously, adapting to a non-stationary stream of environmental stimuli without catastrophically forgetting previously acquired knowledge. In artificial neural networks (ANNs), this stability-plasticity dilemma remains a fundamental challenge \cite{kudithipudi2022biological} (Fig.~ \ref{fig:fig1}a). Standard gradient descent algorithms optimize parameters myopically for immediate performance, overwriting latent representations vital to historical tasks, resulting in a phenomenon known as \textit{catastrophic forgetting} \cite{mccloskey1989catastrophic, french1999catastrophic}. Complementary Learning Systems (CLS) theory proposes that the brain resolves the stability-plasticity dilemma through systems consolidation, in which the hippocampus acts as a rapid buffer for new information that is subsequently integrated into the neocortex \cite{o2014complementary}. This gradual integration is likely governed at the cellular level by metaplasticity, or the "plasticity of synaptic plasticity" \cite{fusi2005cascade, jedlicka2022contributions} (Fig.~ \ref{fig:fig1}b, c).

Deep learning architectures have employed architectural, replay-based, or regularization strategies to facilitate continual learning \cite{wang2024comprehensive}. However, these paradigms frequently diverge from biological constraints \cite{hess2023two}:
\begin{itemize}
    \item \textbf{Architectural and parameter-isolation methods} often suffer from capacity limits or require explicit task boundaries to allocate new subnets.
    \item \textbf{Replay-based frameworks}, while empirically robust, impose substantial memory overhead via episodic buffers and suffer from "stability gaps" at task transition boundaries due to sudden distribution shifts.
    \item \textbf{Regularization techniques} such as Elastic Weight Consolidation (EWC) \cite{kirkpatrick2017overcoming} capture the geometry of the historical loss landscape via the Fisher Information Matrix (FIM). However, these methods typically require explicit task IDs to compute and freeze importance metrics at the end of distinct training phases, relying on additive loss penalties that distort the global optimization landscape.
\end{itemize}

In this work, we argue that synaptic metaplasticity can be mathematically re-framed as a normative, geometry-aware optimization strategy. Rather than retroactively evaluating the importance of parameters, a continual learning agent must treat optimization as navigation through a partially observable Riemannian manifold \cite{vastola2025gradient}. In this space, the local curvature corresponds to the critical structural knowledge of past data. Standard stochastic gradient descent (SGD) operates myopically in Euclidean space. By blindly following the steepest descent of the current stimuli, it violently displaces parameters along directions of varying loss curvatures \cite{kingma2014adam, chng2025preconditioners}. We demonstrate that it can also be used to accelerate convergence by adjusting for these varying loss curvatures \cite{kingma2014adam, chng2025preconditioners}. We demonstrate that it can serve a crucial inverse objective. By mapping the parameter-space geometry of joint tasks in real time, preconditioning can serve as a structural stabilizer for continual learning.

Building upon this geometric intuition, we introduce \textbf{Syn}aptic \textbf{G}eometric \textbf{A}daptive \textbf{P}reconditioning (\textbf{SynGAP}), a gradient-based, task-free continual learning framework. Operating without explicit episodic triggers, SynGAP simulates continuous neuromodulation by maintaining an online exponential moving average of the FIM over a continuous data stream. During each optimization step, this dynamic approximation of the historical loss curvature is mapped to a bounded multiplicative mask. Rather than altering the global loss function with additive penalties, this mask directly preconditions raw stochastic gradients, geometrically attenuating update vectors in directions of high historical curvature while preserving plasticity along flat, non-interfering parameter trajectories.

SynGAP balances the stability-plasticity trade-off by enforcing rigorous mathematical bounds on its preconditioning mask, thereby systematically preventing the loss of plasticity or parameter intransigence that commonly affects rigid regularization methods. By unifying biological metaplasticity with Riemannian parameter space geometry, SynGAP avoids the severe memory footprint of episodic data replay and relies only on a small episodic buffer. Our framework delivers a highly robust, computationally lightweight optimizer for continuous, adaptive intelligence at the edge.

Our core contributions are structured as follows:
\begin{itemize}
    \item We mathematically formalize biological metaplasticity as a gradient preconditioning problem.
    \item We propose the SynGAP optimizer, which continuously tracks parameter importance across non-stationary streams without requiring task boundaries or incurring extensive data storage overhead.
    \item We introduce a bounded multiplicative masking mechanism that explicitly maintains parameter plasticity, providing safeguards against network freezing.
\end{itemize}

\begin{figure*}[htpb!]
\centering
\includegraphics[width=0.75\textwidth]{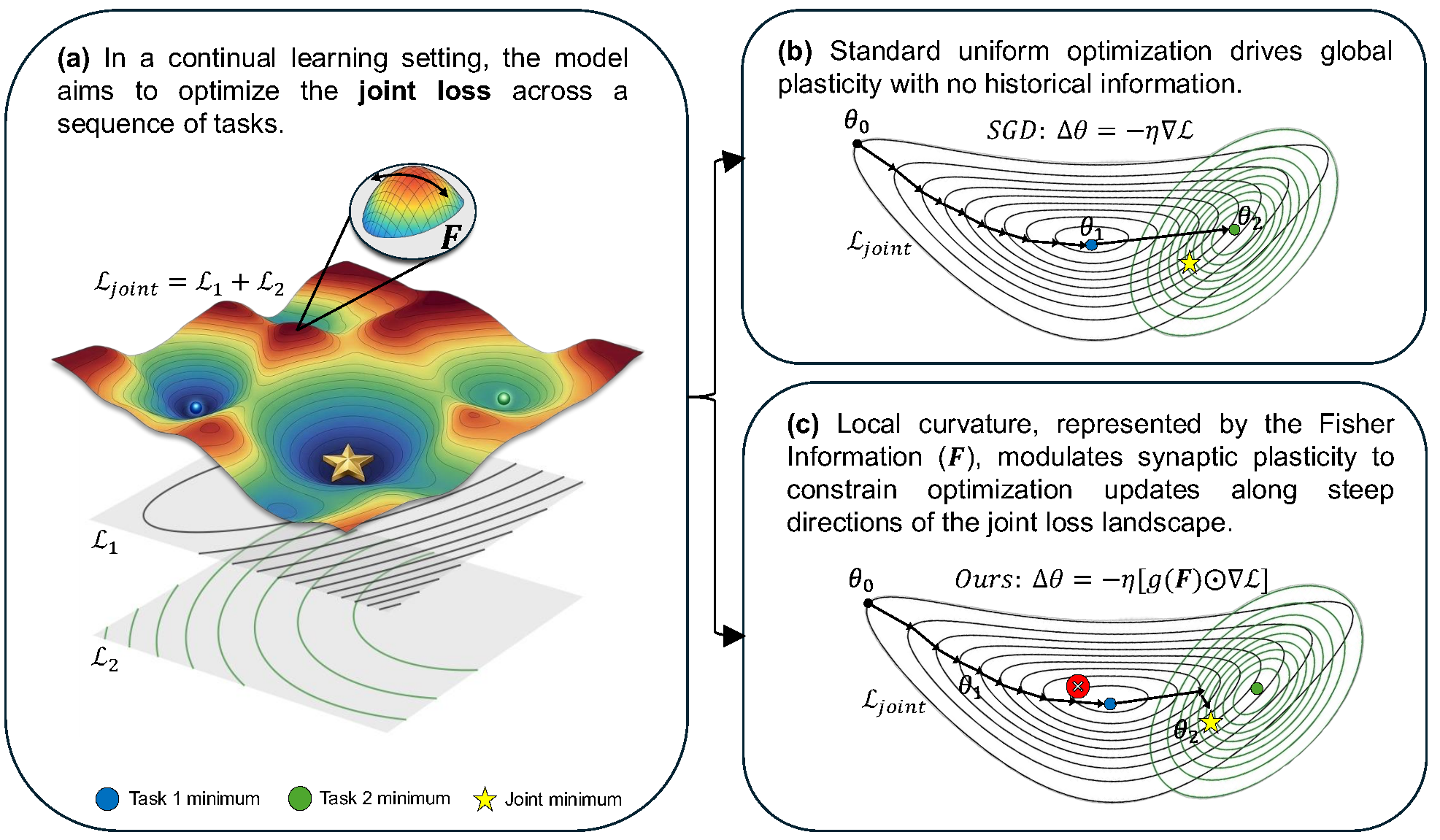}
\caption{\textbf{SynGAP.} \textbf{(a)} To achieve continual learning success, the model must optimize the joint loss of all tasks. \textbf{(b)} Naive optimization struggles to find a joint minimum for multi-task continual learning, as updates are executed globally on the newest task data. \textbf{(c)} \textit{SynGAP} introduces local metaplastic updates that are governed by a function $g(\textbf{F})$ that modulates the optimization step based on Fisher Information of the joint loss. This figure is produced with the aid of Gemini 3.}
\label{fig:fig2} 
\end{figure*} 

\section{Related Works}
\label{sec:relworks}

\subsection{Continual learning {\it via} parameter regularization}
\label{subsec:cl_reg}
A common strategy for mitigating catastrophic forgetting is parameter regularization, which aims to facilitate continual learning by restricting the plasticity of individual weights according to their estimated importance to previously learned tasks. These approaches typically append a penalty term to the global loss function, gently pulling the updated weights back toward their consolidated values. The primary distinction between these methods lies in how they estimate parameter importance. EWC \cite{kirkpatrick2017overcoming} leverages the diagonal of the empirical FIM to measure local parameter sensitivity, but requires strict task IDs and access to true labels. Synaptic Intelligence (SI) \cite{zenke2017continual} continuously tracks a path integral of gradients during training, and Memory-Aware Synapses (MAS) \cite{aljundi2018memory} use the gradient of the network's output function, thereby adding an unsupervised estimate of parameter importance. More recently, Test Time Adaptation-based methods such as Efficient Anti-forgetting Test-time Adaptation (EATA) \cite{niu2022efficient} have extended this paradigm to label-free, streaming environments by employing pseudo-labels to estimate Fisher information and actively filtering high-entropy samples to reduce the computational overhead of backward passes.

Despite their widespread adoption and recent adaptations for efficient inference, regularization methods typically rely on post hoc additive penalties on the loss function, which can cause scaling issues during optimization as the number of tasks increases. By treating consolidation as a static constraint rather than a dynamic property of the active learning process, these methods often suffer from a trade-off between retaining past knowledge and adapting to new data distributions \cite{dohare2024loss}. This limitation underscores the need for approaches that intervene directly in the optimization step, moving away from global loss penalties in favor of embedding memory protection seamlessly into the network's update dynamics.

\subsection{Gradient modification and preconditioning}
\label{subsec:preconditioning}
Moving beyond static loss penalties, gradient preconditioning methods intervene directly in optimization dynamics by scaling weight updates based on historical gradients or loss geometry \cite{kingma2014adam, amari1998natural}. Recent advancements include gradient projection techniques, which update models along orthogonal directions to the old tasks' gradients to minimize interference \cite{farajtabar2020orthogonal}. To mitigate the rigidity of strict orthogonality, Scaled Gradient Projection (SGP) combines orthogonal projections with scaled updates along past core gradient spaces \cite{saha2023continual}. SGP computes this basis importance via Singular Value Decomposition (SVD) on input representations, bypassing the need for data replay \cite{saha2021gradient}.

However, these methods face severe computational bottlenecks, particularly limiting their viability for low-power edge applications. Standard optimizers use transient scaling factors that rapidly overwrite historical importance. Conversely, rigorous projection methods such as SGP require the continuous computation and manipulation of large basis matrices via SVD \cite{saha2021gradient}, incurring prohibitive global overhead. Consequently, the field lacks a localized, permanently stateful mechanism that can embed enduring memory protection directly into gradient updates without expensive mathematical operations. Looking to neuroscience, this gap naturally points toward the complex, history-dependent metaplasticity observed in biological synapses.

\subsection{Biological metaplasticity}
\label{subsec:metaplasticity}
To overcome the limitations of transient optimizers and computationally prohibitive curvature matrices, recent continual learning architectures draw inspiration from the brain's Complementary Learning Systems (CLS) \cite{o2014complementary, pham2021dualnet} and the micro-level dynamics of "complex" synapses. In biological networks, hidden internal states dictate the metaplasticity of a synapse ("the plasticity of its plasticity"), ensuring that prior consolidation structurally preconditions future susceptibility to change \cite{benna2016computational, fusi2005cascade}. Although some continual learning approaches have translated this concept into artificial neural networks \cite{aguilar2025continuous, laborieux2021synaptic}, they are not architecture-agnostic and suffer from a capacity bottleneck that prevents them from capturing complex, high-dimensional distributions or large-scale tasks. Consequently, there is a lack of a continuous, model-agnostic, and highly scalable metaplastic framework.

\section{Preliminaries}
\label{sec:preliminaries}

\paragraph{Continual learning setting.}
In standard task-based continual learning (CL), a model is trained over a sequence of tasks $\mathcal{T} = \{T_1, T_2, \dots, T_T\}$ \cite{mirzadeh2020understanding}. Each task $T_t$ is associated with a specific dataset $\mathcal{D}_t = \{(x_i^t, y_i^t)\}_{i=1}^{N_t}$, where $x_i^t \in \mathcal{X}$ represents an input sample and $y_i^t \in \mathcal{Y}$ is the corresponding label, drawn from a task-specific data distribution $P_t(X, Y)$.

During the training phase for task $T_t$, the model only has access to the current dataset $\mathcal{D}_t$. Access to data from previous tasks is typically restricted to a small memory buffer. The overarching objective after training on $t$ tasks is to find the optimal parameters $\theta$ that minimize the expected risk across all tasks observed so far:
\begin{equation}
    \min_{\theta} \sum_{k=1}^{t} \mathbb{E}_{(x,y) \sim P_k} [\mathcal{L}(f_\theta(x), y)]
\end{equation}
where $f_\theta$ is the neural network parameterized by $\theta$, and $\mathcal{L}$ is a standard loss function (here, we use cross-entropy). The fundamental challenge in this setting is to successfully learn the distribution $P_t$ without severely degrading the performance of previous distributions $P_{1:t-1}$. Minimizing this expected risk is typically done by optimizing the joint loss of all datasets, $\bigcup_{k=1}^t \mathcal{D}_k$, which CL methods must approximate without full access to past data.

\paragraph{The Fisher information matrix.}
To measure the importance of parameters for past data, we rely on the Fisher Information Matrix (FIM) \cite{rattray1998natural, amari1998natural}. The theoretical FIM is defined as the covariance of the score function:
\begin{equation}
    \mathcal{F}(\theta) = \mathbb{E}_{x \sim P_{X}, y \sim P_{\theta}(Y|X)} \left[ \nabla_\theta \log p_\theta(y|x) \nabla_\theta \log p_\theta(y|x)^\top \right]
\end{equation}
Note that the targets $y$ are sampled from the model's predictive distribution $P_{\theta}(Y|X)$, rather than the ground-truth labels (which constitute the Empirical Fisher) \cite{van2025computation}. To maintain computational feasibility, we restrict our formulation to the diagonal approximation of the FIM, denoted as $\textbf{F} \in \mathbb{R}^{|\theta|}$ \cite{soen2024trade}.

\paragraph{Gradient preconditioning.}
Standard gradient descent updates parameters via $\theta_{t+1} = \theta_t - \eta \nabla_\theta \mathcal{L}_t$. Preconditioned gradient methods modify the update direction by applying a transformation matrix $P^{-1}$:
\begin{equation}
    \theta_{t+1} = \theta_t - \eta P^{-1} \nabla_\theta \mathcal{L}_t
\end{equation}
While adaptive optimizers like Adam \cite{kingma2014adam} construct $P^{-1}$ to accelerate optimization in high-curvature directions, our framework leverages preconditioning in the opposite direction to attenuate updates in directions critical to preserving historical knowledge.

\section{Methodology}
\label{sec:methodology}

In this section, we introduce SynGAP. Let $\theta \in \mathbb{R}^D$ denote the flattened parameter vector of the neural network and $\mathcal{L}(\theta)$ denote the empirical loss function over the continuous data stream.

\paragraph{Online Fisher information tracking.}
Unlike methods that compute importance metrics at the task boundary, SynGAP maintains a continuous estimate of parameter importance. We simulate the accumulation of the metaplastic state by computing an online Exponential Moving Average (EMA) of the diagonal FIM. The global state vector $\textbf{F}$ is updated continuously:
\begin{equation}
    F_{t} = \alpha F_{t-1} + (1 - \alpha) \text{diag}(\mathcal{F}_{t}(\theta))
\end{equation}
where $\alpha \in [0, 1)$ is the metaplastic retention rate, and $\mathcal{F}_{t}(\theta)$ is the instantaneous Fisher information evaluated on the current batch.

\paragraph{Bounded metaplastic preconditioning.}
Raw FIM values can vary by orders of magnitude, contributing to instability in optimization. To translate the unbounded metaplastic state vector $F$ into a stable learning mechanism, we introduce a globally normalized plasticity function, denoted $g(\textbf{F})$.

Let $\mu = \frac{1}{D} \sum_{i=1}^D F_i$ be the global mean of the Fisher EMA across all network parameters. We define a dynamic scaling factor $c$ to normalize the landscape: 
\begin{equation}
    c = \frac{\text{arctanh}(\tau)}{\mu + \lambda}
\end{equation}
where $\tau \in (0, 1)$ is the target consolidation hyperparameter.

We define the plasticity function as:
\begin{equation}
    g(\textbf{F}) = 1 - \tanh(c(\textbf{F} + \lambda)) + \epsilon
    \label{eq:plasticity_func}
\end{equation}

where $\lambda$ acts as a Tikhonov damping term for numerical stability, and $\epsilon$ is a small scalar representing ambient plasticity (to prevent complete intransigence). The output of this function yields the consolidation mask, $\Omega = g(\textbf{F})$, which is strictly bounded within $[\epsilon, 1+\epsilon]$. All operations in Eq.~\ref{eq:plasticity_func} are applied element-wise.

As illustrated in Fig.~\ref{fig:fig2}, the SynGAP architecture operates via two parallel streams during the optimization step. While the standard backward pass computes the raw task gradients $\nabla_\theta \mathcal{L}(\theta)$, the metaplastic tracking module updates simultaneously the historical Fisher EMA state $F$. 

The plasticity function $g(F)$ acts as a gating mechanism, transforming the unbounded historical state $F$ into the bounded multiplicative mask $\Omega$. The raw gradients are then preconditioned by this mask before altering the network. The final SynGAP update rule at optimization step $t$ is formulated as:
\begin{equation}
    \theta_{t+1} = \theta_t - \eta (g(\textbf{F}) \odot \nabla_\theta \mathcal{L}(\theta))
    \label{eq:SynGAP_update}
\end{equation}
where $\eta$ is the global learning rate and $\odot$ denotes the element-wise Hadamard product. By mathematically decoupling the gradient computation from the metaplastic gating, we demonstrate that SynGAP requires no additive loss penalty, cleanly avoiding the catastrophic gradient scaling and loss-landscape distortion issues common in prior continual learning methods \cite{li2023fixed}.

\section{Experiments}
\label{sec:experiments}

To validate the efficacy of SynGAP, we evaluate our framework on challenging benchmarks for class-incremental and continuous domain-incremental learning. We aim to answer three key questions: (1) How does SynGAP compare to state-of-the-art continual learning methods, (2) How does metaplastic preconditioning compare to standard adaptive optimizers, and (3) What is the impact of the bounded plasticity mask?

\begin{table*}[ht!]
    \centering
    \caption{Performance metrics on Split CIFAR-100 and CORe50. Average Accuracy (ACC), Average Forgetting (FM), and Intransigence (INT) are reported in $\%$. \textit{Joint (oracle)} serves as an upper bound for accuracy. All methods are averaged over five full runs. Best results are in bold.}
    \resizebox{\textwidth}{!}{%
    \begin{tabular}{lcccccc}
        \toprule
        & \multicolumn{3}{c}{Split CIFAR-100} & \multicolumn{3}{c}{CORe50} \\
        \cmidrule(lr){2-4} \cmidrule(lr){5-7}
        Method & ACC ($\uparrow$) & FM ($\downarrow$) & INT ($\downarrow$) & ACC ($\uparrow$) & FM ($\downarrow$) & INT ($\downarrow$) \\
        \midrule
        Joint (oracle) & $67.76 \pm 0.00$ & $--$ & $--$ & $99.60 \pm 0.00$ & $--$ & $--$ \\
        \midrule
        SGD & $6.60 \pm 1.50$ & $51.64 \pm 1.26$ & $12.12 \pm 2.32 $ & $58.97 \pm 0.61$ & $41.56 \pm 0.79$ & $1.88 \pm 0.12$ \\
        SGDM & $8.77 \pm 0.07$ & $71.98 \pm 1.85$ & $\bf{-5.90 \pm 1.87}$ & $58.41 \pm 1.04$ & $43.37 \pm 1.28$ & $1.08 \pm 0.06$ \\
        Adam & $7.59 \pm 0.26$ & $72.61 \pm 0.43$ & $-5.20 \pm 0.35$ & $58.02 \pm 0.67$ & $47.24 \pm 0.79$ & $\bf{-0.04 \pm 0.05}$ \\
        Ballistic & $7.43 \pm 0.75$ & $63.04 \pm 1.19$ & $1.38 \pm 0.63$ & $55.34 \pm 1.18$ & $49.79 \pm 1.28$ & $0.15 \pm 0.04$ \\
        ER & $17.60 \pm 4.25$ & $48.56 \pm 3.32$ & $5.23 \pm 2.22
        $ & $67.22 \pm 2.00$ & $33.72 \pm 2.43$ & $3.02 \pm 0.57$ \\
        EWC++ & $6.80 \pm 0.87$ & $64.36 \pm 1.60$ & $2.33 \pm 2.43$ & $40.51 \pm 1.69$ & $60.53 \pm 0.51$ & $5.67 \pm 1.18$ \\
        \textbf{SynGAP (Ours)}& $\bf{27.28 \pm 0.58}$ & $\bf{37.87 \pm 2.98}$ & $6.91 \pm 3.88$ & $\bf{68.65 \pm 1.71}$ & $\bf{30.85 \pm 1.61}$ & $4.09 \pm 0.62$ \\
        \bottomrule
    \end{tabular}
           }
    \label{tab:main_results}
\end{table*}

\subsection{Experimental setup}
\paragraph{Benchmarks.} We evaluate on two standard datasets:
\begin{itemize}
    \item \textit{Split CIFAR-100:} A class-incremental benchmark of $100$ classes, split into $10$ tasks (with $10$ classes per task). Although the dataset contains discrete shifts, SynGAP processes the stream continuously without access to task identifiers or boundary signals.
    \item \textit{CORe50:} A continuous object recognition benchmark designed for highly non-stationary environments, comprising $50$ classes. We utilize the New Classes (NC) scenario, which mimics a smooth, continuous data stream \cite{lomonaco2017core50}.
\end{itemize}

\paragraph{Baselines.} We compare SynGAP with two categories of baselines:
\begin{itemize}
    \item \textit{Continual Learning Baselines:} Experience Replay (ER) \cite{rolnick2019experience} and the online version of Elastic Weight Consolidation, EWC++ \cite{kirkpatrick2017overcoming, chaudhry2018riemannian}. All strategies operate without requiring task IDs. All sampling buffers used are equal to $20n$ for $n$ classes.
    \item \textit{Optimizer Baselines:} SGD, SGD with Momentum (SGDM), Adam, and Ballistic \cite{vastola2025gradient}.
\end{itemize}

\paragraph{Evaluation Metrics.} To assess performance, we report the Average Accuracy (\textit{ACC}) \cite{lopez2017gradient}, where the mean performance across the data stream is measured after the model is completely trained. Let $ACC \in [0, 100]$ as a percentage value.

\begin{equation}
\text{ACC} = \frac{1}{T} \sum_{i=1}^{T} R_{T,i}
\label{eq:acc}
\end{equation}

In Equ.~\ref{eq:acc}, $R_{i,j}$ is the test classification accuracy of the model on task $T_j$ after observing the last sample from task $T_i$. For additional continual learning metrics, we also report Average Forgetting (\textit{FM}) and Intransigence (\textit{INT}), as first introduced in \cite{chaudhry2018riemannian}. Let $FM \in [-100, 100]$ and $INT \in [-100, 100]$ be in percent. Specifically, \textit{FM} evaluates catastrophic forgetting by measuring the average drop in accuracy for each task from its peak performance to its final performance at the end of training. 



Conversely, \textit{INT} measures the model's resistance to acquiring new information, calculated as the performance gap between a specific model and a reference upper-bound model (here, we use one trained jointly on all data). 



Together, these metrics provide a holistic assessment of the balance between the stability and plasticity of the model capacity \cite{jedlicka2022contributions}. All evaluation metrics are presented as percentage values.

\subsection{Main results}
\label{subsec:results}
Table~\ref{tab:main_results} summarizes the performance of our proposed method, SynGAP, against several baselines on the Split CIFAR-100 and CORe50 benchmarks. We evaluate the models based on Average Accuracy (ACC), Average Forgetting (FM), and Intransigence (INT). The \textit{Joint (oracle)} training performance is provided as an upper bound.

\paragraph{Superior Accuracy and Forgetting Mitigation}

SynGAP significantly outperforms all baseline methods in both overall accuracy and memory retention across both datasets. 

On the challenging Split CIFAR-100 benchmark, standard optimization strategies (SGD, SGDM, Adam, and Ballistic), along with the regularization-based EWC++ method, suffer from severe catastrophic forgetting, yielding final accuracies below $9\%$ and forgetting measures exceeding $51\%$. While ER provides a strong baseline by boosting accuracy to $17.60\%$, SynGAP establishes superior accuracy of $27.28\%$ and reduces average forgetting to $37.87\%$. 

Similar trends are observed with the CORe50 dataset. SynGAP achieves the highest accuracy at $68.65\%$, outperforming the ER baseline ($67.22\%$). EWC++ notably struggles on this continuous benchmark, achieving a much lower accuracy of 40.51\% and a high forgetting measure of 60.53\%. Furthermore, SynGAP demonstrates the highest resistance to catastrophic forgetting, with an FM of $30.85\%$, compared to ER's $33.72\%$ and standard SGD's $41.56\%$.

\paragraph{The intransigence trade-off}

While SynGAP achieves the best ACC and FM, the Intransigence (INT) metric reveals an expected trade-off inherent to continual learning systems. 
Methods utilizing momentum or adaptive learning rates (SGDM and Adam) achieve negative intransigence scores on Split CIFAR-100 ($-5.90\%$ and $-5.20\%$, respectively) and near-zero scores on CORe50. This indicates high plasticity and the ability to learn new tasks rapidly. However, this comes at the direct cost of catastrophic forgetting, as evidenced by their high FM scores and extremely low average accuracies on CIFAR-100. 

Regularization approaches like EWC++ can struggle with this balance, sometimes becoming overly rigid—as seen by its high INT score of 5.67\% on CORe50. In contrast, SynGAP maintains a stable INT score ($6.91\%$ on CIFAR-100 and $4.09\%$ on CORe50), which is only a few percentage points higher than those of the other methods. This demonstrates that SynGAP successfully restricts excessive parameter updates to protect historical knowledge (i.e., lowering FM) without becoming overly rigid or severely impeding the model's ability to learn incoming tasks.

\subsection{Optimization trajectory}

\begin{figure}[]
\centering
\includegraphics[width=0.50\textwidth]{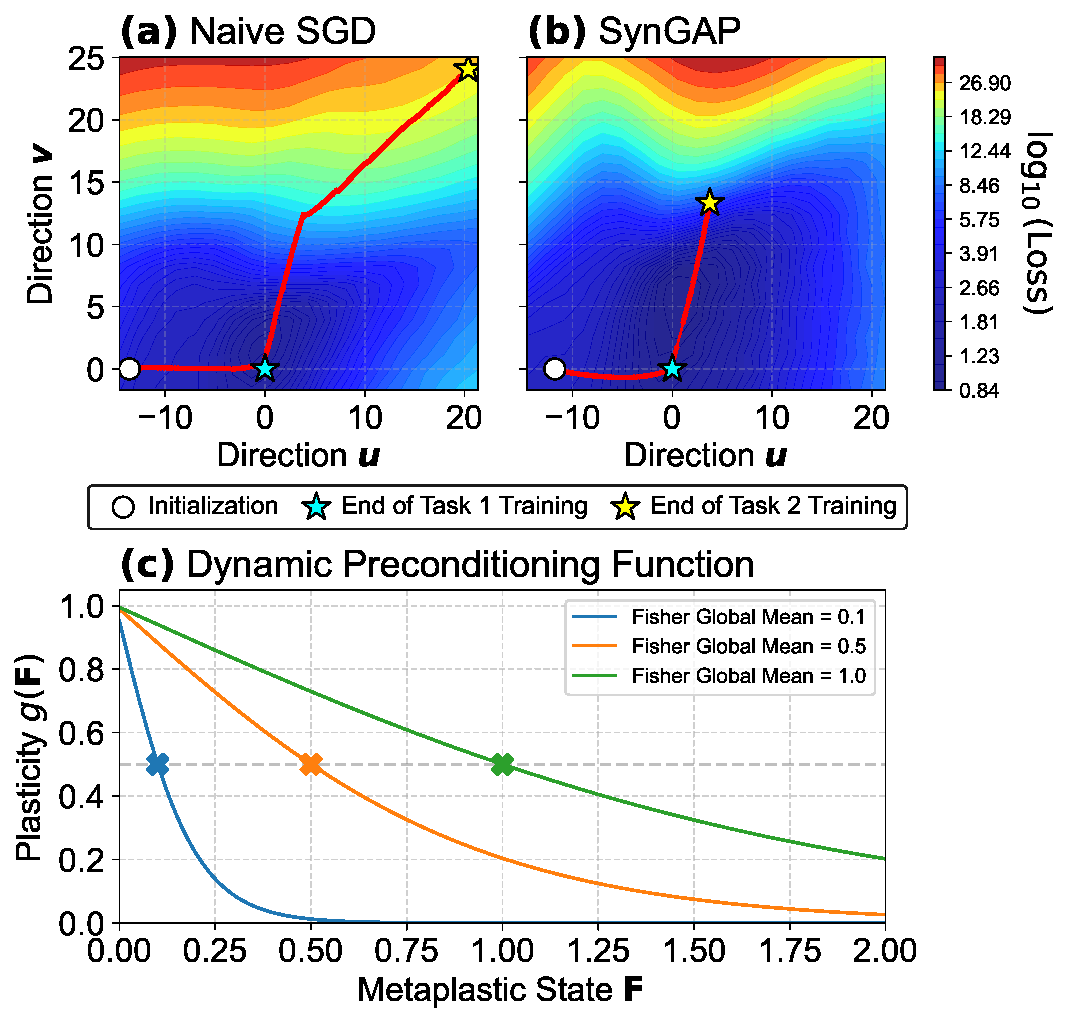}
\caption{\textbf{SynGAP smooths the optimization trajectory compared to standard optimizers.} On the initial task loss landscape, the optimization trajectory from initialization (white circle) to task 1 training (cyan star) to task 2 training (yellow star) is shown using \textbf{(a)} SGD, and \textbf{(b)} our method, SynGAP. The proposed method navigates towards a wider, more stable basin between task 1 and task 2. \textbf{(c)} The dynamic preconditioning function, as described in Equ.~\ref{eq:plasticity_func} is applied in SynGAP to modulate the gradient steps shown in \textit{(b)}. Plasticity is dictated by the metaplastic state as defined by Fisher.}
\label{fig:fig3} 
\end{figure} 

\begin{table*}[htbp!]
    \centering
    \caption{Ablation study of core components and hyperparameter sensitivity on Split CIFAR-100. Performance is evaluated across Average Accuracy (ACC), Forgetting Measure (FM), and Intransigence (INT) in $\%$. Each configuration is averaged over five full experimental runs.}
    \label{tab:ablation}
    \begin{tabular}{lccc}
        \toprule
        \textbf{Configuration} & \textbf{ACC} $\uparrow$ & \textbf{FM} $\downarrow$ & \textbf{INT} $\downarrow$ \\
        \midrule
        Full Model & $27.28 \pm 0.58$ & $37.87 \pm 2.98$ & $6.91 \pm 3.88$ \\
        \midrule
        \multicolumn{4}{@{}l}{\textbf{Component Ablations}} \\
        \quad w/ Unbounded Linear Penalty (aka w/o Mask) & $14.72 \pm 0.34$ & $49.46 \pm 2.39$ & $8.18 \pm 2.66$ \\
        \quad w/o Metaplastic Preconditioning (aka ER) & $17.60 \pm 4.25$ & $54.23 \pm 0.53$ & $5.23 \pm 2.22$ \\
        \midrule
        \multicolumn{4}{@{}l}{\textbf{Metaplastic Retention Rate ($\alpha$)}} \\
        \quad Low $\alpha$ (Rapid Plasticity) & $19.27 \pm 1.18$ & $46.32 \pm 6.84$ & $7.39 \pm 5.54$ \\
        \bottomrule
    \end{tabular}
\end{table*}

To better understand how our approach prevents catastrophic forgetting, we visualize the model's optimization trajectory in Fig.~\ref{fig:fig3}. The figure projects the loss landscape onto a two-dimensional plane defined by the direction vectors $\mathbf{u}$ and $\mathbf{v}$. Following the visualization techniques in \cite{li2018visualizing}, we use filter-normalized random directions to plot the loss contours of Task 1 of the Split CIFAR-100 setting. As illustrated in Fig.~\ref{fig:fig3}, the dotted lines trace the optimization path of the model parameters ($\theta$) as it learns Task 2, starting from model initialization.

As shown in Fig.~\ref{fig:fig3}(a), the Naive SGD trajectory exhibits clear signs of catastrophic forgetting. Although the model successfully locates a low-loss minimum at the end of Task 1, the subsequent training phase for Task 2 drives the parameters into a region of significantly higher loss, demonstrating the destruction of previously acquired representations.

However, our method visualized in Fig.~\ref{fig:fig3}(b) shows how the trajectory during Task 2 is visibly restricted, ensuring that the final parameters remain anchored within a broader low-loss basin associated with Task 1. 

This constrained trajectory is directly driven by the function detailed in Fig.~\ref{fig:fig3}(c), which depicts the dynamic preconditioning function. The plasticity term $g(\mathbf{F})$ parameter-wise gates on the optimization steps, decaying as the metaplastic state $\mathbf{F}$ increases. The decay rate is dynamically adjusted based on the global mean of the Fisher information. Through this plasticity function, SynGAP heavily penalizes movement along dimensions critical to historical tasks while maintaining high plasticity in directions where $\mathbf{F}$ remains low, naturally guiding the optimization trajectory along the contours of the previous task's loss landscape.

\subsection{Ablation studies}
To isolate the contributions of the SynGAP method, we perform the following ablations:
\begin{itemize}
    \item \textbf{The Impact of the Bounded Mask:} We replace the $\tanh$ bounded plasticity function $g(\bf{F})$ with an unbounded linear penalty (similar to standard regularization). 
    \item \textbf{Ablation of Preconditioning:} To isolate the effect of metaplastic preconditioning, we evaluate the full model against an ablated variant where preconditioning is removed, effectively reducing the system to standard ER.
    \item \textbf{Metaplastic Retention Rate ($\alpha$):} We analyze the sensitivity of the system by decreasing the EMA decay parameter $\alpha$, inducing extreme plasticity.
\end{itemize}

Table~\ref{tab:ablation} presents the empirical results of these configurations on the Split CIFAR-100 benchmark.

First, substituting the bounded plasticity mask with an unbounded linear penalty results in severe performance degradation. The average accuracy (ACC) drops significantly from 27.28\% (Full Model) to 14.72\%, while the forgetting measure (FM) increases from 37.87\% to 49.46\%. Intransigence (INT) also rises slightly to 8.18\%. This confirms that the globally normalized $\tanh$ mapping is essential for preventing gradient-scaling instability and effectively shielding consolidated parameters from destructive updates.

Second, removing the metaplastic preconditioning entirely, which reduces the architecture to a standard ER baseline, results in a substantial increase in forgetting. Without preconditioning, the FM spikes to 54.23\% and overall accuracy falls to 17.60\%. Although this ablation yields a lower INT of 5.23\%, indicating greater plasticity, it comes at a greater cost to memory retention. This validates the use of preconditioning and its role in retaining historical knowledge.

Finally, lowering the metaplastic retention rate ($\alpha$) induces plasticity, causing the model to prematurely overwrite the historical importance of its parameters. This high-plasticity regime destabilizes the learning process, reducing the final accuracy to 19.27\% and worsening the forgetting measure to 46.32\%. This shows that a slower decay rate (i.e., deep consolidation) is necessary to ensure the model retains long-term structural knowledge while sequentially adapting to new distributions.


\section{Discussion}
\label{sec:discussion}

The empirical success of SynGAP highlights a fundamental flaw in traditional parameter-regularization techniques for CL. Methods like EWC rely on additive loss penalties, which inherently alter the global optimization landscape and often lead to gradients that scale disproportionately with the number of tasks. By shifting to a strictly multiplicative preconditioning paradigm bounded by $\tanh$, SynGAP guarantees that the loss landscape of the current stream remains undistorted, while structurally critical weights are protected via gradient attenuation rather than loss manipulation. Furthermore, replacing task-based Fisher calculations with an online EMA allows SynGAP to organically track non-stationary distributions, bridging the gap between CL regularization and standard adaptive optimization.

Beyond correcting the optimization landscape, SynGAP exhibits a powerful synergy with experience replay. Standard Experience Replay (ER) often struggles with gradient interference, where updates from the memory buffer conflict with the high-magnitude gradients of the novel data stream. By preconditioning the gradients, SynGAP acts as a geometric gatekeeper, ensuring that updates driven by new data are projected away from the critical dimensions of previously learned tasks. This alignment reduces catastrophic interference at the synaptic level, explaining the significant performance gap between SynGAP and standard ER observed in our experiments.

Additionally, the task-free nature of SynGAP offers a distinct operational advantage. Traditional consolidation methods require discrete task boundaries to compute and anchor the importance matrices, rendering them unsuitable for continuous, real-world data streams. By integrating the Fisher estimation directly into the online training step, SynGAP achieves continuous consolidation with minimal computational overhead. It eliminates the need for expensive offline Fisher re-computations and complex regularization-scaling heuristics.

\paragraph{Limitations.} 
Despite its robustness, the SynGAP framework operates under several assumptions that present avenues for future work. 
First, to maintain computational efficiency, SynGAP relies on the diagonal approximation of the Fisher Information Matrix (FIM). While sufficient, this ignores off-diagonal dependencies. In highly complex loss landscapes, a block-diagonal or Kronecker-factored (K-FAC) approximation may be required to capture true geometric constraints \cite{martens2015optimizing}. 
Second, while metaplastic preconditioning is memory-efficient, SynGAP still relies on a sampling buffer to provide the active-learning signal. 
Finally, the hyperparameter $\alpha$ (the metaplastic retention rate) assumes a uniform timescale of forgetting across all layers, which may not accurately reflect the abstraction of hierarchical features in deep networks \cite{ly2025optimization, erhan2009difficulty}.

\paragraph{Future directions.} 
Future work will explore hierarchical timescales to more accurately simulate localized biological metaplasticity \cite{behrouz2025nested}. Additionally, extending the scalar plasticity mask $\Omega$ to incorporate second-order off-diagonal geometric constraints could eliminate the need for the ambient plasticity term $\epsilon$, allowing for true parameter lockdown without risking dead networks.

\section{Conclusion}
\label{sec:conclusion}
In this work, we introduced SynGAP, a task-free Continual Learning framework that re-conceptualizes biological metaplasticity as online geometric gradient preconditioning. By maintaining a continuous exponential moving average of the Fisher Information Matrix and projecting it through a globally normalized bounding function, SynGAP selectively attenuates gradients to protect historical knowledge without requiring explicit task boundaries or additive loss penalties. Our empirical evaluations demonstrate that SynGAP successfully prevents the catastrophic forgetting exhibited by standard adaptive optimizers (such as Adam) on non-stationary streams, while substantially outperforming traditional regularization and replay-based baselines. By mathematically formalizing the biological concept of synaptic consolidation into a stable, multiplicative optimization rule, SynGAP offers a highly robust solution for continuous, adaptive intelligence.

\subsection*{Acknowledgements}
The authors acknowledge the support from the Australian Research Council under Project DP230100019. I.A. acknowledges support from the Australian Government’s Research Training Program (RTP).

\subsection*{Code Availability} \label{sec:codest}
The code generated during this study are available from the corresponding author upon reasonable request.

\subsection*{Data Availability} \label{sec:datast}
We employed publicly accessible datasets that can be accessed through the following links:
\begin{itemize}
  \item[---] CIFAR-100: \url{https://www.cs.toronto.edu/~kriz/cifar.html}, and 
  \item[---] CORe50:  \url{https://vlomonaco.github.io/core50}.
\end{itemize}

\bibliography{literature}


\end{document}